\documentclass[letterpaper, 10 pt, conference]{ieeeconf}  

\IEEEoverridecommandlockouts                              

\usepackage{graphics} 
\usepackage{epsfig} 
\usepackage{mathptmx} 
\usepackage{times} 
\usepackage{amsmath} 
\usepackage{amssymb}  

\usepackage{hyperref} 
\usepackage{xcolor} 
\usepackage{multirow} 

\title{\LARGE \bf KITTI-CARLA: a KITTI-like dataset generated by CARLA Simulator}

\author{Jean-Emmanuel Deschaud$^{1}$

\thanks{$^{1}$ MINES ParisTech, PSL University, Centre for Robotics, 75006 Paris, France
\{jean-emmanuel.deschaud@mines-paristech.fr\}}

}

\begin{document}

\maketitle
\thispagestyle{empty}
\pagestyle{empty}

\begin{abstract}

KITTI-CARLA is a dataset built from the CARLA v0.9.10 simulator~\cite{dosovitskiy2017carla} using a vehicle with sensors identical to the KITTI dataset~\cite{geiger2012kitti}. The vehicle thus has a Velodyne HDL64 LiDAR positioned in the middle of the roof and two color cameras similar to Point Grey Flea 2. The positions of the LiDAR and cameras are the same as the setup used in KITTI.

The objective of this dataset is to test approaches of semantic segmentation LiDAR and/or images, odometry LiDAR and/or image in synthetic data and to compare with the results obtained on real data like KITTI. This dataset thus makes it possible to improve transfer learning methods from a synthetic dataset to a real dataset.

We created 7 sequences with 5000 frames in each sequence in the 7 maps of CARLA providing different environments (city, suburban area, mountain, rural area, highway...). The dataset is available at~\url{http://npm3d.fr}

\end{abstract}

\section{INTRODUCTION}

Our approach is to generate a synthetic dataset in a realistic simulated environment and having the ground truth for semantic segmentation, instance segmentation, odometry poses.

\section{CARLA}

\begin{figure}[ht]
    \centering
    \includegraphics[width=0.8\linewidth]{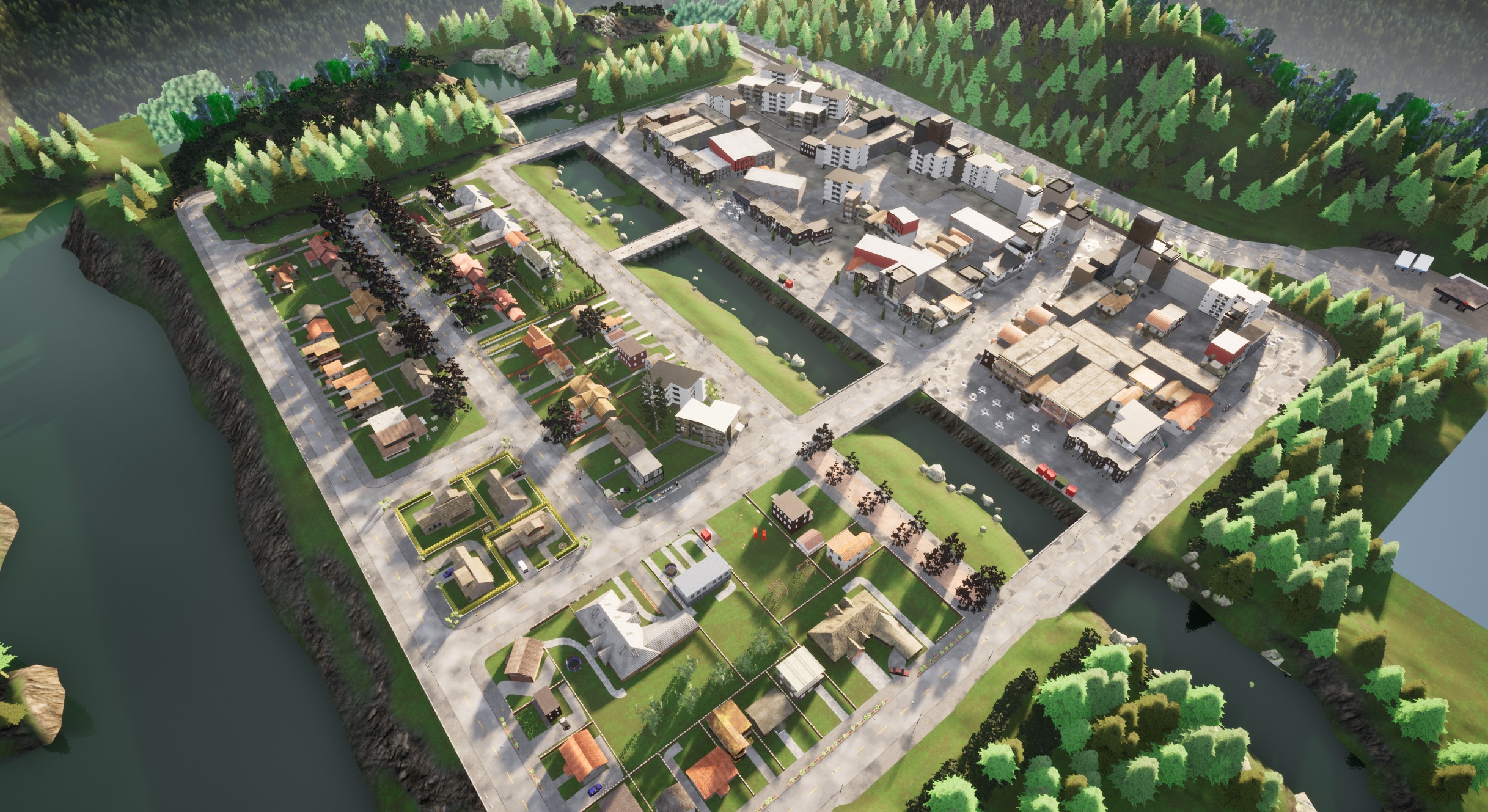}
    \includegraphics[width=0.8\linewidth]{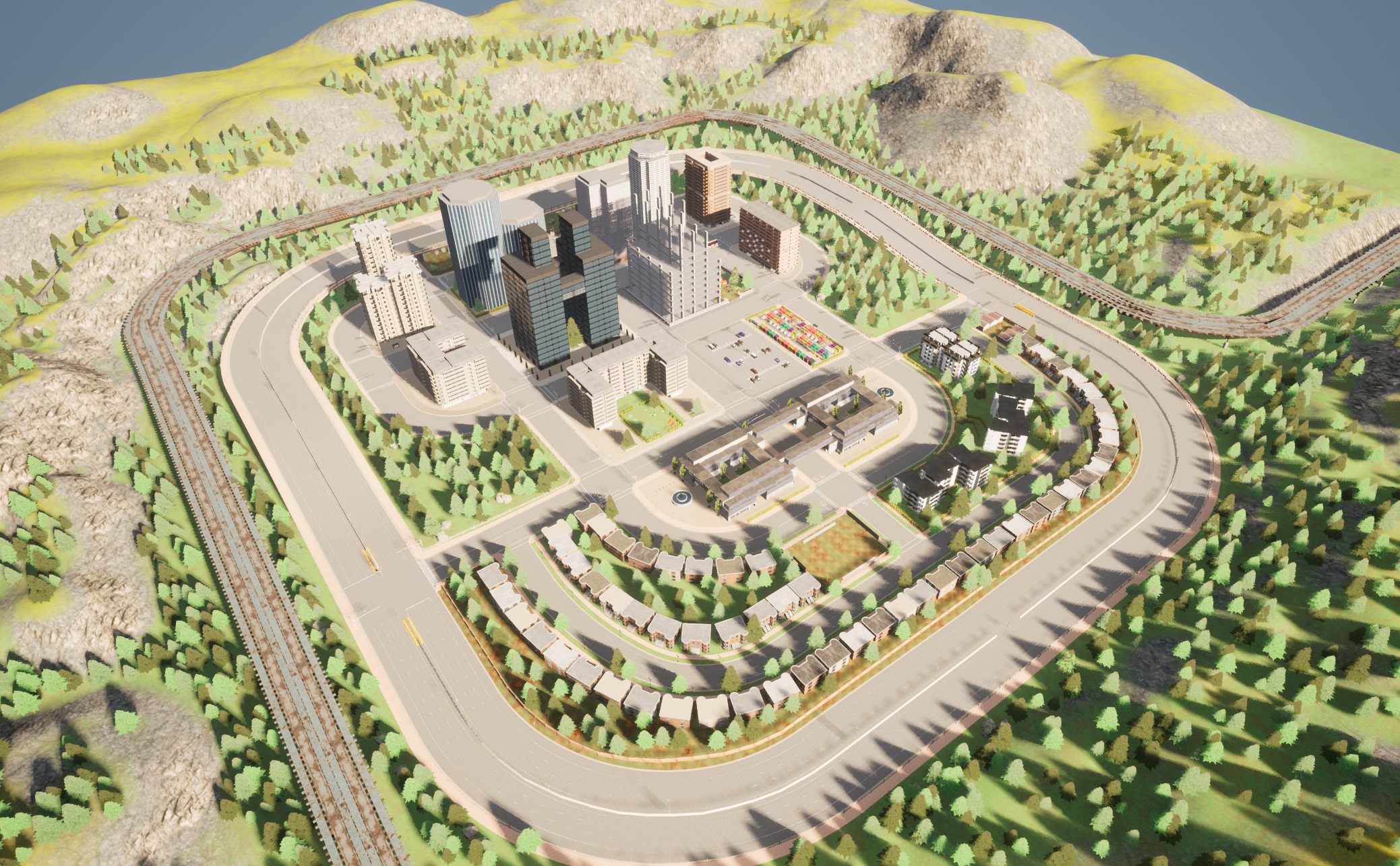}
    \includegraphics[width=0.8\linewidth]{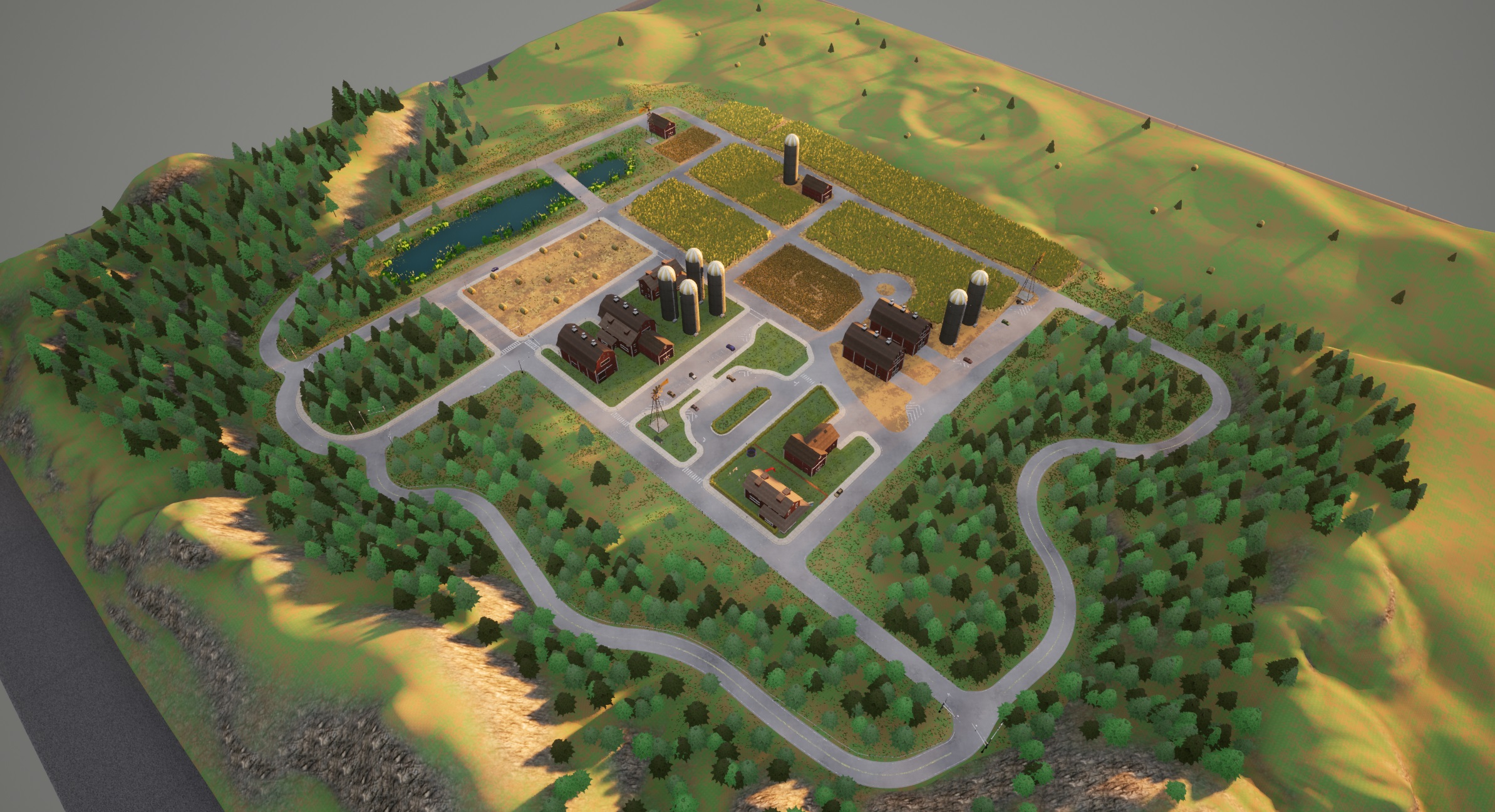}
    \caption{Examples of maps from the CARLA Simulator used to create our synthetic dataset. Respectively from top to bottom: city (Town01), highway (Town05) and farm (Town07).}
    \label{maps}
\end{figure}

CARLA is an open source simulator for autonomous vehicles which allows to simulate multiple sensors, especially cameras and LiDARs (Figure~\ref{maps}). We use it to simulate a LiDAR and two color cameras boarded on the roof of a car traveling randomly in accordance with the traffic laws. The simulation is launched with a fixed time step of 0.001s. This means that the complete physics will be updated at every step: we generate all the data at each step, then move on to the next step until the end of the simulation. This frequency of 1000Hz makes it possible to simulate the rolling-shutter effect of LiDAR sensors.\\

\section{SETUP}

\begin{figure}[ht]
    \centering
    \includegraphics[width=0.8\linewidth]{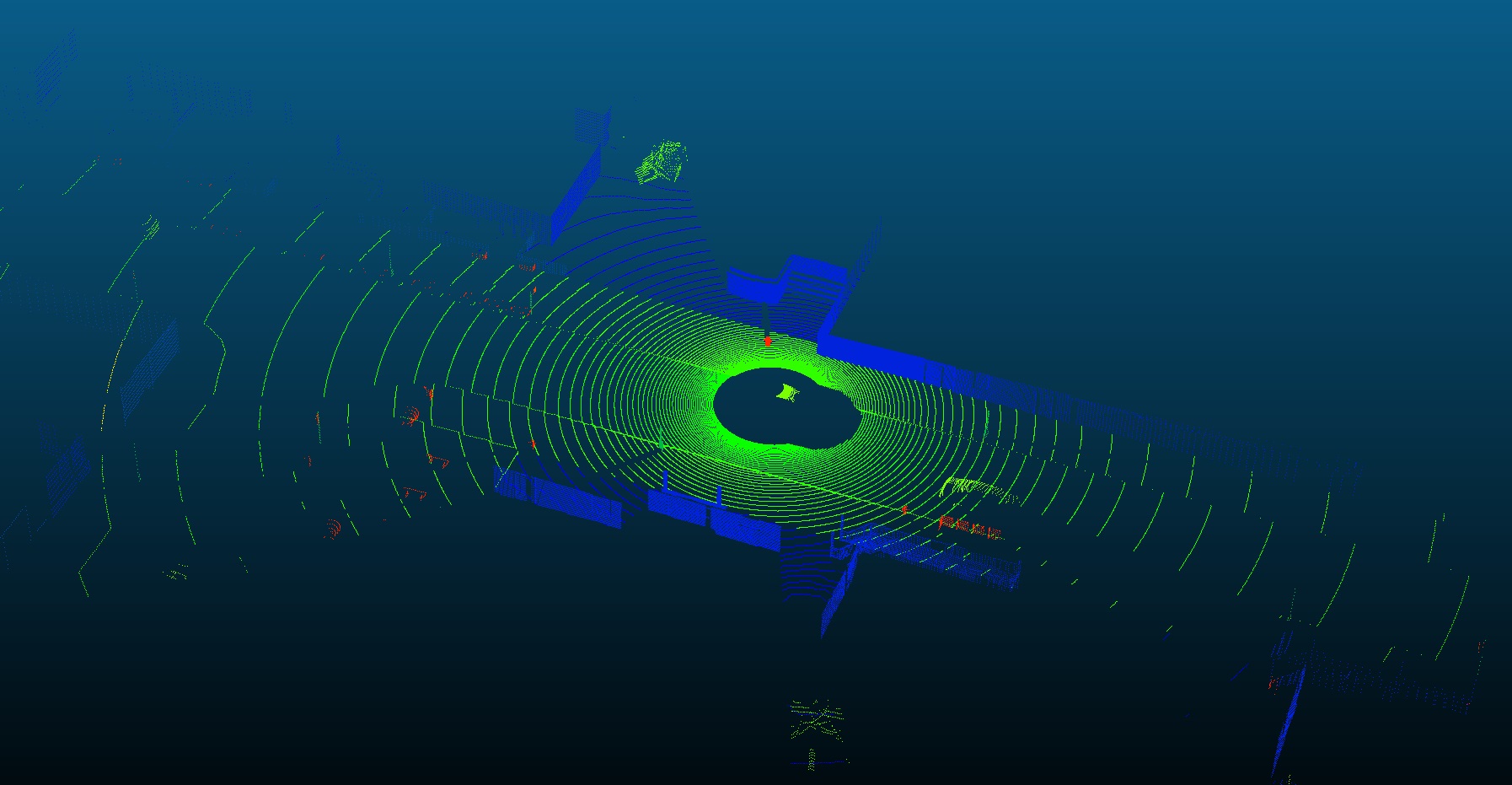}
    \includegraphics[width=0.8\linewidth]{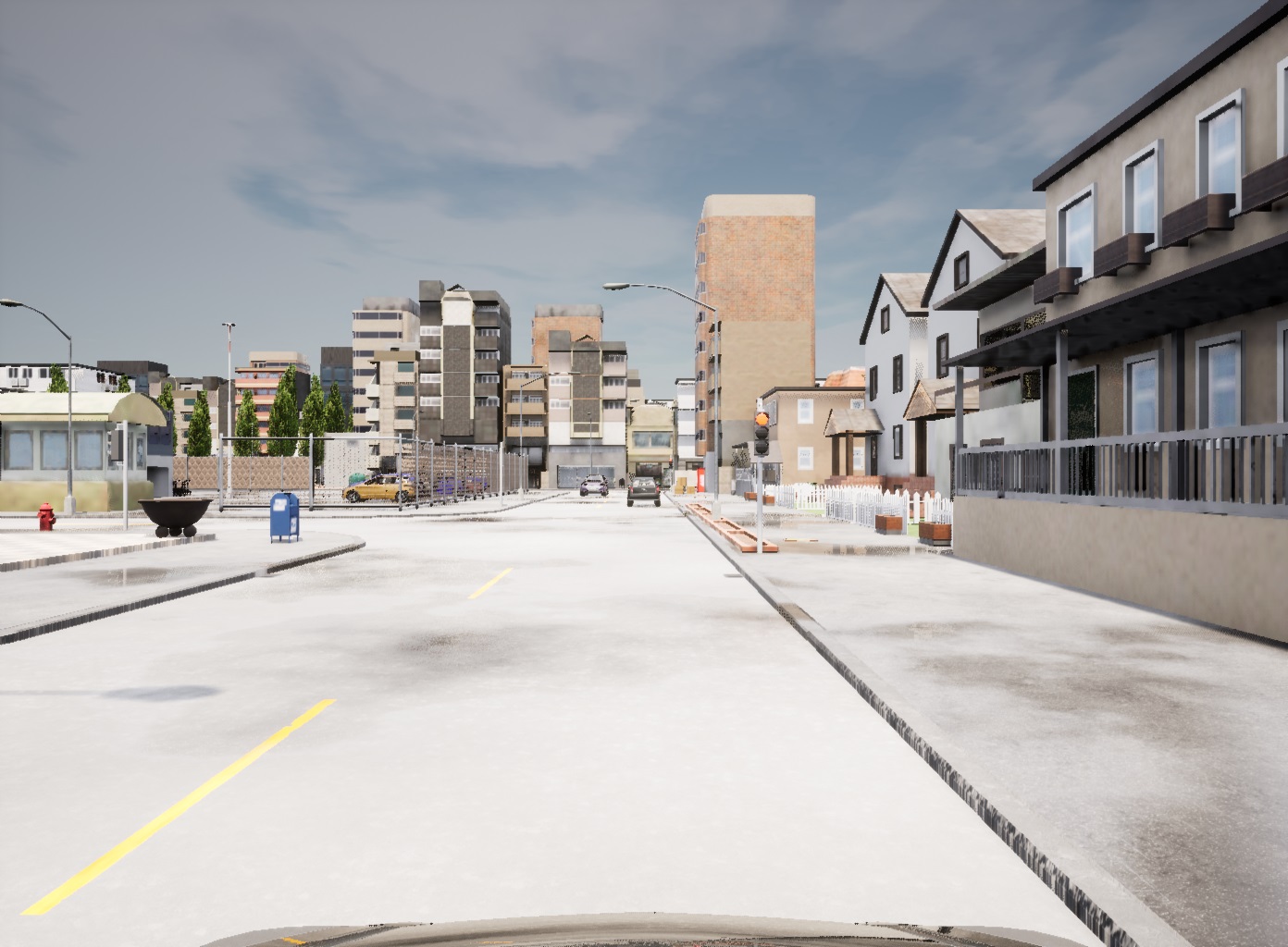}
    \caption{Top, one frame of the LiDAR with ground truth semantic segmentation in color. Bottom, an image from the camera 0.}
    \label{view_lidar_camera}
\end{figure}

The simulated LiDAR is vertical and mounted on top of the roof of the vehicule. It has the same characteristics as a Velodyne HDL-64E (64 channels, from $-24.8^{\circ}$ to $+2^{\circ}$ vertical field of view, 10Hz revolution frequency, 80m maximum range so around 130k points per frame). The LiDAR is attached to the vehicle body with a global translation of ($X=0.0m$, $Y=0.0m$, $Z=1.80m$) and is pointing backwards (Figure~\ref{view_lidar_camera}).\\

Let's define a frame as a complete revolution of the data coming from the LiDAR. During the revolution, the LiDAR attached to the vehicle moves. Because LiDAR frequency is 10Hz and CARLA simulation is 1000Hz, at each simulator step the LiDAR returns 1/100 of a frame. During this interval, the physics is not updated so all the points in a measurement reflect the same "static picture" of the scene taken at a same time.\\

We provide with the dataset all the poses of the LiDAR at 1000Hz allowing to know the ground truth of the poses to generate a point cloud. We provide a python script to compute the position of each point in world coordinate (thanks to the LiDAR transformation saved at each step).\\

The two simulated cameras have a resolution of 1392x1024 pixels with 72 degrees FOV. They work at 10Hz. Cameras are attached to the vehicle body with a global translation of ($X=0.30m$, $Y=0.0m$, $Z=1.70m$) for camera 0 and ($X=0.30m$, $Y=0.50m$, $Z=1.70m$) for camera 1 (Figure~\ref{view_lidar_camera}).\\

\section{DATASET}

We recorded 7 sequences using the 7 maps from Town01 to Town07 (Figure~\ref{maps}). We generated 5000 frames per sequence and for each of them we saved the full LiDAR transformations in world coordinates and corresponding timestamps in a ASCII file. We also saved the timestamps of the camera images and the calibration between the LiDAR and the 2 cameras in ASCII files.\\

\bibliographystyle{IEEEtran}
\bibliography{IEEEabrv,biblio}

\begin{thebibliography}{1}
\providecommand{\url}[1]{#1}
\csname url@rmstyle\endcsname
\providecommand{\newblock}{\relax}
\providecommand{\bibinfo}[2]{#2}
\providecommand\BIBentrySTDinterwordspacing{\spaceskip=0pt\relax}
\providecommand\BIBentryALTinterwordstretchfactor{4}
\providecommand\BIBentryALTinterwordspacing{\spaceskip=\fontdimen2\font plus
\BIBentryALTinterwordstretchfactor\fontdimen3\font minus
  \fontdimen4\font\relax}
\providecommand\BIBforeignlanguage[2]{{%
\expandafter\ifx\csname l@#1\endcsname\relax
\typeout{** WARNING: IEEEtran.bst: No hyphenation pattern has been}%
\typeout{** loaded for the language `#1'. Using the pattern for}%
\typeout{** the default language instead.}%
\else
\language=\csname l@#1\endcsname
\fi
#2}}

\bibitem{dosovitskiy2017carla}
\BIBentryALTinterwordspacing
A.~Dosovitskiy, G.~Ros, F.~Codevilla, A.~Lopez, and V.~Koltun, ``{CARLA}: {An}
  open urban driving simulator,'' in \emph{Proceedings of the 1st Annual
  Conference on Robot Learning}, ser. Proceedings of Machine Learning Research,
  S.~Levine, V.~Vanhoucke, and K.~Goldberg, Eds., vol.~78.\hskip 1em plus 0.5em
  minus 0.4em\relax PMLR, 13--15 Nov 2017, pp. 1--16. [Online]. Available:
  \url{http://proceedings.mlr.press/v78/dosovitskiy17a.html}
\BIBentrySTDinterwordspacing

\bibitem{geiger2012kitti}
A.~Geiger, P.~Lenz, and R.~Urtasun, ``Are we ready for autonomous driving? the
  kitti vision benchmark suite,'' in \emph{2012 IEEE Conference on Computer
  Vision and Pattern Recognition}, 2012, pp. 3354--3361.

\end{thebibliography}

\end{document}